\documentclass{article}

\PassOptionsToPackage{numbers, compress}{natbib}
\usepackage[preprint]{neurips_2024}

\usepackage[utf8]{inputenc} % allow utf-8 input
\usepackage[T1]{fontenc}    % use 8-bit T1 fonts
\usepackage{hyperref}       % hyperlinks
\usepackage{url}            % simple URL typesetting
\usepackage{booktabs}       % professional-quality tables
\usepackage{amsfonts}       % blackboard math symbols
\usepackage{nicefrac}       % compact symbols for 1/2, etc.
\usepackage{microtype}      % microtypography
\usepackage[dvipsnames]{xcolor}         % colors
\usepackage{graphicx}
\usepackage[accsupp]{axessibility}
\usepackage{orcidlink}
\usepackage{todonotes}
\usepackage{multirow}
\usepackage{pifont}% http://ctan.org/pkg/pifont
\usepackage{colortbl}
\usepackage{xspace}
\usepackage{amsmath}
\usepackage{amssymb}
\usepackage{mathtools}
\usepackage{pgfplots}
\pgfplotsset{compat=1.18} 
\usepackage{algorithm}
\usepackage{algpseudocode}
\usepackage[capitalize,noabbrev]{cleveref}

\usepackage{amsmath,amsfonts,bm}

\def\eqref#1{equation~\ref{#1}}
\def\1{\bm{1}}

\DeclareMathAlphabet{\mathsfit}{\encodingdefault}{\sfdefault}{m}{sl}
\SetMathAlphabet{\mathsfit}{bold}{\encodingdefault}{\sfdefault}{bx}{n}

\DeclareMathOperator*{\argmin}{arg\,min}

\crefname{figure}{Fig.}{Figs.}
\crefname{equation}{Eq.}{Eqs.}

\newcommand{\cmark}{\ding{51}}
\newcommand{\xmark}{\ding{55}}
\newcommand{\ccmark}{\textcolor{Green}{\ding{51}}}
\newcommand{\cxmark}{\textcolor{red}{\ding{55}}}
\newcommand{\ours}{AptDet\xspace}

\title{Aligned Unsupervised Pretraining of Object Detectors with Self-training}

\author{
Ioannis Maniadis Metaxas$^{1}$\thanks{Corresponding author: i.maniadismetaxas@qmul.ac.uk} \quad 
Adrian Bulat$^{2}$ \\
\textbf{Ioannis Patras}$^1$ \quad 
\textbf{Brais Martinez}$^2$ \quad 
\textbf{Georgios Tzimiropoulos}$^{1,2}$ \\
$^1$Queen Mary University of London \quad $^2$Samsung AI Cambridge
}

\begin{document}

\maketitle

\begin{abstract}
The unsupervised pretraining of object detectors has recently become a key component of object detector training, as it leads to improved performance and faster convergence during the supervised fine-tuning stage. 
Existing unsupervised pretraining methods, however, typically rely on low-level information to define proposals that are used to train the detector. Furthermore, in the absence of class labels for these proposals, an auxiliary loss is used to add high-level semantics. 
This results in complex pipelines and a task gap between the pretraining and the downstream task.
We propose a framework that mitigates this issue and consists of three simple yet key ingredients:
   (i) richer initial proposals that do encode high-level semantics,
   (ii) class pseudo-labeling through clustering, that enables pretraining using a standard object detection training pipeline,
   (iii) self-training to iteratively improve and enrich the object proposals.
Once the pretraining and downstream tasks are aligned, a simple detection pipeline without further bells and whistles can be directly used for pretraining and, in fact, results in state-of-the-art performance on both the full and low data regimes, across detector architectures and datasets, by significant margins. 
We further show that our pretraining strategy is \textit{also} capable of pretraining from scratch (including the backbone) and works on complex images like COCO, paving the path for unsupervised representation learning using object detection directly as a pretext task.

\end{abstract}

\section{Introduction} 
\label{sec:intro}

Object detection has been a major challenge in computer vision and the focus of extensive research efforts.
Two complementary avenues of research have led to several breakthroughs: 
a) more powerful architectures, such as the end-to-end single stage DETR~\cite{detr} family of detectors, and 
b) unsupervised detector pretraining, which leverages unlabeled data to improve performance on downstream tasks where annotations are expensive, ambiguous, and/or imprecise.
Notably, existing pretraining methods largely focus on DETR detectors, as they are the current state-of-the-art, but show slow training convergence and are sample inefficient (i.e. require large amounts of annotated data).

Typically, unsupervised detector pretraining methods generate object proposals (bounding boxes or segmentation masks) randomly~\cite{dai2021up}, through heuristic-based methods~\cite{bar2022detreg}, or using unsupervised localization techniques~\cite{wang2023cut}. The pretraining task is then to localize the proposals in the image and to distinguish object vs no-object regions. Thus, while the downstream task (detection) requires \textit{both} the localization and the classification of the objects, proposals are typically generated using low-level information and neglect class-level information. To add discrimination based on high-level semantics, current methods typically add a second auxiliary loss, often a variant of contrastive learning. As a result, their pipelines often involve student-teacher models, feature-matching objectives, aggressive color \& cropping augmentations, and other complex and computationally costly mechanisms. Despite these efforts, the two aspects (localization and discrimination) are not adequately integrated, leading to a task gap with the downstream task. Consequently, most current detector pretraining methods suffer large performance degradation when unfreezing the backbone, highlighting the task misalignment problem and preventing the end-to-end joint pretraining of the detector head and the backbone. 

 \begin{figure}[t]
   \begin{center}
    \includegraphics[width=0.8\linewidth]{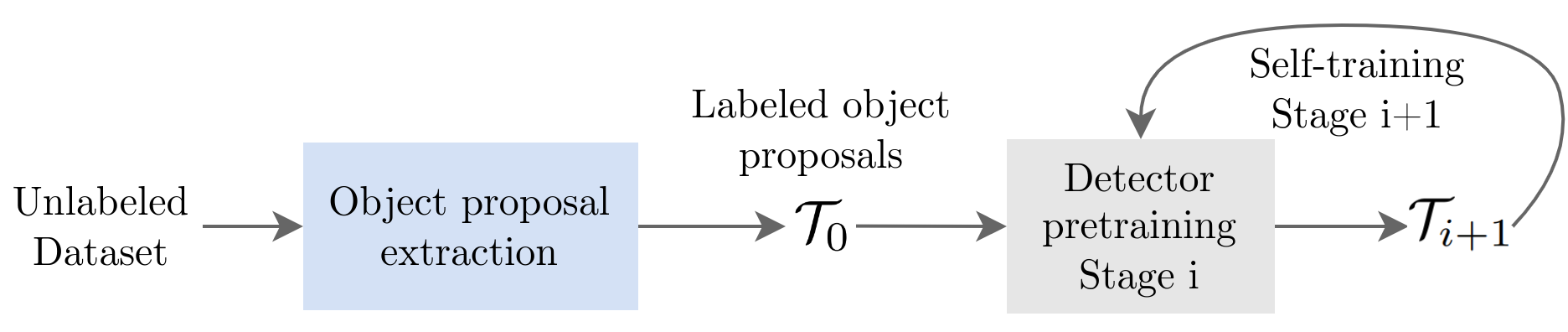}
    \caption{\textbf{\ours overview:} \textbf{(i)} Object proposals are extracted from images in an unsupervised manner and assigned pseudo-labels via clustering; \textbf{(ii)} The pseudo-labeled object proposals are used to train the detector, which learns to localize objects and discriminate their pseudo-class label; \textbf{(iii)} The detector then generates a new set of proposals and pseudo-labels, which are used for self-training. %Best seen in color.
    }
    \vspace{-0.5cm}
    \label{fig:main_fig}
    \end{center}
 \end{figure}

This work proposes \ours (Aligned PreTraining for Detection), 
a novel detector pretraining framework that solves the aforementioned limitations, achieves state-of-the-art performance across benchmarks, and 
can effectively train the entire detector architecture (backbone and detector head) jointly, and even from scratch, and on scene-centric images, e.g. COCO or OpenImages. 
Our method, seen in~\cref{fig:main_fig}, has three main components, which we show \textit{are all needed} for highly effective pretraining:
\\
\textbf{(i) Unsupervised proposal extraction from feature maps:} We obtain proposals based on high-level semantic information by clustering feature maps produced by a self-supervised backbone. We avg-pool the resulting masks to obtain an embedding for each proposal that captures high-level semantics.
\\
\textbf{(ii) Detector pretraining with pseudo-labels:} The per-proposal high-level semantic embeddings are then clustered across the dataset, and cluster membership is used as pseudo-class labels. We then use the proposals and pseudo-class labels as training data in a standard object detection training pipeline.
\\
\textbf{(iii) Iterative self-training:} We observe that the detector resulting from component (ii) can produce better proposals than the ones it was trained on. We find that detection pretraining can be applied in an iterative fashion, where the current pretrained model produces the pseudo-labels to train itself further with improved supervision.

We conduct extensive experiments with several detector architectures and report two main findings:
\\
\textbf{(1) Improved detection \& segmentation accuracy:} \ours consistently outperforms previous works by significant margins, across architectures, and benchmarks for unsupervised detector pretraining.
\\
\textbf{(2) Self-supervised representation learning from complex images:} We show that \ours can effectively train the whole network (detector head and backbone jointly) from scratch directly, and on complex images, demonstrating impressive flexibility for unsupervised representation learning.

\section{Related Works} \label{sec:related_works}

\textbf{Unsupervised object detector pretraining:}
Object detector pretraining methods aim to pretrain the detector architecture, in addition to the backbone. Previous work in this area has mostly focused on DETR detectors, which can achieve great performance but exhibit sample inefficiency and slow convergence. 
Thus, detector pretraining (as opposed to backbone-only pretraining) is an important task for such methods.
Among these, UP-DETR~\cite{dai2021up} proposed randomly selecting areas from each image, extracting feature representations, and injecting them into the DETR detector's queries. The detector was then trained to localize the areas to which the injected representations corresponded.
DETReg~\cite{bar2022detreg} subsequently used Selective Search~\cite{uijlings2013selective} to generate object proposals as annotations for the detector. The detector was trained both to localize the proposals and represent them, mimicking a pretrained backbone encoder. 
JoinDet~\cite{joindet} improved upon DETReg by replacing Selective Search with a dynamic object proposal method that inferred the location of objects from the detector's internal activations. 
Siamese DETR~\cite{huang2023siamese} used instead a student-teacher multi-view architecture for pretraining where, in addition to class-agnostic localization, the detector is trained to learn transformation-invariant representations at the global (image) and local (object) level. 
Finally, SeqCo-DETR~\cite{jin2023seqco} proposed sequence consistency as a pretext task, combined with a masking strategy.
Notably, most of these works freeze the detector's backbone encoder during pretraining, as they suffer performance drops otherwise~\cite{dai2021up,bar2022detreg}. This is a significant limitation, as it prevents true end-to-end self-supervised training, and makes such frameworks heavily dependent on the quality of the pretrained backbone.
Beyond DETR-focused works, AlignDet~\cite{li2023aligndet} focused on pretraining on the smaller COCO dataset using contrastive learning.
Importantly, all of these works uniformly pretrain detectors in \textit{a class-unaware manner} and rely on auxiliary objectives to improve the detectors' discriminative capacity. This creates a misalignment between the pretraining task and the downstream task of class-aware object detection, which limits the pretraining's effectiveness. We emphasize that this also applies to AlignDet, which uses the term alignment to describe the fact that the detector is pretrained and fine-tuned on the same dataset (COCO). This is entirely distinct from our approach of explicitly aligning the objectives of the pretext and downstream tasks. Notably, when compared with AlignDet, we show that \ours outperforms it by a significant margin.

\textbf{Unsupervised backbone pretraining for dense prediction:} Most works on unsupervised pretraining focus on pretraining the network backbone, rather than the full object detection network\cite{pixpro,detcon,soco,vangansbeke2021revisiting,densecl,huang2022learning,gokul2022refine,detco,slotcon,odin,bai2022point,karlsson2021vice,islam2023self,ding2022deeply,li2022univip,xie2021unsupervised}.
Specifically, works in this area do not include a localization component (i.e. they do not localize objects in images) and typically only pretrain the backbone focusing solely on representation learning. They are, therefore, distinct from unsupervised detector pretraining works, which train the detector and include a localization task, while often using pretrained backbones as initialization.

\textbf{Unsupervised object localization:} Different from object detector pretraining, this task aims to localize all objects in an image in an unsupervised manner, without considering class information~\cite{van2022discovering,simeoni2021localizing,wang2022tokencut,simeoni2022unsupervised,melas2022deep,wang2022freesolo}.
Among these works, CutLER~\cite{wang2023cut} and FreeSOLO~\cite{wang2022freesolo} are notable for also using self-training.
We emphasize that the main goal of these works is object localization/discovery, not the training of powerful detectors. 
Accordingly, the detectors trained by these works typically are not evaluated by finetuning with annotated data. 
Such methods also typically restrict their proposals to the most confident few (often just one) to avoid false positives, which is not well suited for detector pretraining, where training benefits from a rich set of object proposals covering as many objects (or object parts) as possible, not only the few most prominent ones.
We validate this in our experiments, where we outperform the state-of-the-art in unsupervised object localization~\cite{wang2023cut}.

\textit{Summary of differences with previous works:} 
We outline the features of~\ours relative to previous works in~\cref{tab:distinctions}. 
\ours fundamentally differs from previous works in the following respects:
%\\
\textbf{i)} Whereas previous works define better auxiliary training tasks, we move away from this paradigm and instead align the pretraining and downstream tasks. This simplifies the training process and improves performance.
%\\
\textbf{ii)} Our proposal extraction and self-training approach produces object proposals that are based on high-level semantic information and are rich and varied, in order to train a powerful detector. That is in contrast, in terms of both implementation and motivation, to works such as CutLER, where proposals are aggressively filtered to a very few confident candidates to promote accurate localization, both in the extraction and self-training stages.
%\\
\textbf{iii)} \ours is designed to facilitate end-to-end detector pretraining, including the backbone, and we demonstrate its effectiveness across architectures and even with from-scratch pretraining. Instead, prior works almost entirely focus on fine-tuning the detection head while keeping the backbone frozen.
 
\setlength{\tabcolsep}{1pt}

\begin{table}[t]
    \caption{Key distinctions among unsupervised detector pretraining methods.}
    \label{tab:distinctions}
  \begin{center}
  \scriptsize{
  \begin{tabular}
  {
  c
  >{\centering\arraybackslash}m{1.9cm}
  >{\centering\arraybackslash}m{1.9cm}
  >{\centering\arraybackslash}m{1.9cm}
  >{\centering\arraybackslash}m{2.5cm}
  >{\centering\arraybackslash}m{1.9cm}
  }
    \toprule
    Method & Downstream Task Alignment & Self-Training & Backbone Pre-training & Proposals from high-level semantics & Rich Object Proposals \\
     \midrule
     AlignDet~\cite{li2023aligndet} & \cxmark & \cxmark & \cxmark & \cxmark & \ccmark \\
     FreeSOLO~\cite{li2023aligndet} & \cxmark & \ccmark & \ccmark & \ccmark & \cxmark \\
     CutLER~\cite{wang2023cut} & \cxmark & \ccmark & \ccmark & \ccmark & \cxmark \\
     UP-DETR~\cite{dai2021up} & \cxmark & \cxmark & \cxmark & \cxmark & \ccmark \\
     DETReg~\cite{bar2022detreg} & \cxmark & \cxmark & \cxmark & \cxmark & \ccmark \\
     JoinDet~\cite{joindet} & \cxmark & \cxmark & \cxmark & \ccmark & \ccmark \\
     SeqCo-DETR~\cite{jin2023seqco} & \cxmark & \cxmark & \ccmark & \cxmark & \ccmark \\
     Siamese DETR~\cite{huang2023siamese} & \cxmark & \cxmark & \cxmark & \cxmark & \ccmark \\
     \arrayrulecolor{lightgray}
    \textbf{\ours} & \ccmark & \ccmark & \ccmark & \ccmark & \ccmark \\
         \arrayrulecolor{black}
    \bottomrule
  \end{tabular}
  }
  \end{center}
    \vspace{-0.4cm}
\end{table}

\setlength{\tabcolsep}{10pt}

\section{Method} \label{sec:method}
 
\ours aims to simplify and better align the pretraining with respect to the downstream task (class-aware detection). To this end, we produce object proposals in the form of \textit{bounding box and pseudo-class label} pairs in an unsupervised manner and then employ a self-training strategy to pretrain and iteratively refine the detector.

 \begin{figure}[t]
   \begin{center}
    \includegraphics[width=0.8\linewidth]{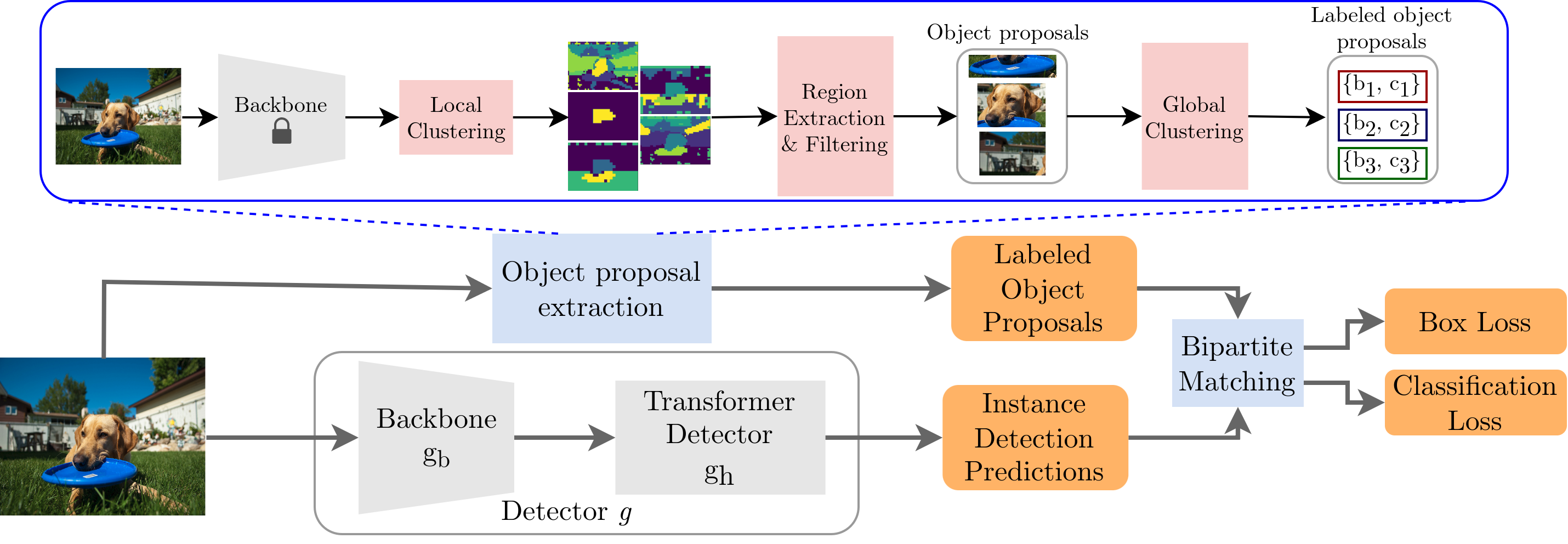}
    \caption{Overview of \ours's pretraining Stage 1 for a DETR-based detector. Pseudo-labeled region proposals are extracted at the start of training, leveraging a self-supervised pretrained backbone. Those proposals are then used to train the detector to both localize objects within the image, and to discriminate their pseudo-labels. 
    \vspace{-0.5cm}
    }
    \label{fig:stage_1}
    \end{center}
 \end{figure}

\subsection{Improved object proposals}
\label{sec:initialization}

Existing works either generate a very limited initial set of proposals to facilitate high precision, or use methods like Selective Search~\cite{uijlings2013selective} that can generate many proposals by relying on low-level priors such as color and texture. Both are suboptimal, the former due to the weaker supervisory signal and the latter because it does not capture high-level semantics. Our aim is to address this gap by utilizing semantic information from self-supervised image encoders to produce rich object proposals and coherent pseudo-class labels.
Specifically, we extract feature maps using a pretrained self-supervised encoder and leverage a bi-level clustering strategy. The first level (termed local clustering) results in bounding box proposals and associated feature representations. The second level, termed global clustering, uses cluster membership to assign a pseudo-class label to each proposal.
Our method leads to rich and diverse region proposals and is essential for the state-of-the-art results of \ours, which we discuss in detail in the ablations presented in~\cref{sec:ablations}.

\textbf{Unsupervised proposal extraction:} Given an input image $X\in \mathbb{R}^{3\times H\times W}$, we use a self-supervised pretrained encoder to extract feature maps $\mathbf{F}_l\in \mathbb{R}^{d_l\times H_l\times W_l}$ from each of the encoder's levels $l$.
Given a feature map $\mathbf{F}$, we employ pixel-wise clustering to group semantically similar features (local clustering). This results in a set of masks $\mathbf{M}=\{\mathbf{m}_k\}_{k=1:K}$, where $K$ represents the number of clusters, which is a user-defined parameter. To provide good coverage for all objects in the image, we apply clustering with different values $K \in \mathcal{K}$ and use feature maps from different layers $l \in\mathcal{L}$, leading to a set of masks $\mathbb{M}=\bigcup \{ \mathbf{M}^{l,K}\}_{K\in\mathcal{K},l\in\mathcal{L}}$. 
Next, the different connected components of each mask are computed, leading to a set of regions $\mathbb{R}$. Each region $\mathbf{r} \in \mathbb{R}$ is then used to extract a bounding box (proposal) $b$ and a corresponding feature vector $f$, where $f$ is computed by average-pooling the last layer feature map $\mathbf{F}_L$ over $\mathbf{r}$. 

\textbf{Proposal filtering:} Due to the clustering at multiple levels of the encoder, the process leads to noisy and overlapping proposals. We employ a number of filters to refine them, such as merging proposals that have a high IoU and proposals with highly related semantic content. This results in a set of $N(i)$ bounding box-feature vector pairs for image $i$, $\{ b_n,f_n\}_{n=1}^{N(i)}$.

\textbf{Pseudo-class label generation:} We then cluster proposals across the whole dataset (global clustering) based on the feature vectors, i.e. we perform a single clustering round on $\{f_{n}^i\}_{n=1:N(i)}^{i=1:I}$, obtaining clusters $S_c$ for $c\in \{1,..,\mathcal{C}\}$. This results in a training set $\mathcal{T}_0=\{X_i, \{(b^i_n,c^i_n)\} \}$, where $c^i_n$ is defined by cluster membership, i.e. $f^i_n \in S_{c^i_n}$.
We use Spectral Clustering~\cite{ng2001spectral} for local and K-Means for global clustering in order to facilitate scaling to large datasets: while Spectral Clustering typically performs better, K-Means is much more efficient and can handle billions of data points~\cite{billionscalesearch}. Therefore, we chose K-Means over alternative clustering algorithms to present and evaluate \ours in its most scalable form.
We note, however, that any clustering algorithm may be used in either case.

\subsection{Pretraining and Self-Training}
\label{sec:self-training}

We can now use the training set $\mathcal{T}_0$ to train an object detector. 
In particular, given an input image and its corresponding extracted object proposals $y$, the network predicts a set $\mathbf{\hat{y}}=\{ \hat{y}_{q} \}_{q=1}^{Q}$, where $\hat{y}_q=( \hat{b}_q, \hat{c}_q )$ comprises the predicted bounding box and predicted category. We note that the extracted proposals $y$ are padded to size $Q$ with $\varnothing$ (no object). We emphasize that \ours is compatible with \textit{any detector architecture}, as we train the detector on simple class-aware detection.
Here, for ease of notation and without loss of generality, we assume a DETR-based detector. The ground truth and the predictions are put in correspondence via bipartite matching, formally defined in~\cref{eq:bipartite}, where $\mathfrak{S}_Q$ is the space of permutations of $Q$ elements. The loss between $\mathbf{y}$ and $\mathbf{\hat{y}}$ is computed in~\cref{eq:detr_loss}, as a combination of a bounding box matching loss and a class matching loss:

\begin{equation}\label{eq:bipartite}
\hat{\sigma} = \argmin_{\sigma \in \mathfrak{S}_Q}
\end{equation}

\vspace{-0.2cm}

\begin{equation}
\sum_{q=1}^Q\left(-log \hat{p}_{\hat{\sigma}(q)} (c_q)+ \text{ \large $\mathbf{1}$}_{\{c_q \neq \varnothing\}} \mathcal{L}_{box}(b_q,\hat{b}_{\hat{\sigma}(q)}) \right)
\label{eq:detr_loss},
\end{equation}

where $\hat{p}$ indicates the predicted per-class probabilities. The indicator function $\text{ \large $\mathbf{1}$}_{c_i \neq \emptyset}$ represents that the box loss only applies to predictions that have been matched to object proposals $y$. Minimizing this loss results in weights $\Theta_0$.

Upon training the detector in this way, we observe that it can identify more objects than those in our original proposals. Critically, this includes smaller and more challenging objects, which contribute to a stronger supervisory signal. We thus generate a new set of pseudo-labels for image $i$ as $\{g(X_i;\Theta_0)\}$, where $g=(g_b,g_h)$ are the detection network, backbone and head respectively.
Previous methods that leverage self-training~\cite{wang2022freesolo,wang2023cut} filter proposals through a confidence threshold. \ours deviates from this strategy: we consider the top-100 proposals of the detector and filter them only in terms of overlap, so that any two boxes have an IOU lower than 0.55 (following~\cite{solovyev2021weighted}), with only the most confident box being kept when such conflicts exist.
We make this choice for two reasons. Firstly, it allows for straightforward application across architectures: different detector architectures/sizes have different behaviors in terms of prediction confidence, which would require tuning the threshold for each case. Secondly, our approach provides better supervision for the pretraining task: we find that a confidence threshold leads to the removal of challenging instances (e.g. small, partially occluded, or uncommon objects). While previous methods focused on object localization and therefore sought to minimize false positives, our goal is to train a strong detector, which requires the supervisory signal of challenging instances.
After this filtering is applied, the result is a new training set $\mathcal{T}_1$.
A new set of weights $\Theta_i$ can be obtained by using the training set $\mathcal{T}_i$ and using $\Theta_{i-1}$ to initialize the weights. Simultaneously, $\Theta_i$ can be used to generate a new training set $\mathcal{T}_{i+1}$. While this process can be iterated indefinitely, we notice optimal performance involves just two rounds of training, which we refer to as Stages 1 \& 2. Stage 1 training, including the proposal extraction process for $\mathcal{T}_{0}$ is shown in~\cref{fig:stage_1}.
The whole method is summarized in~\cref{alg:method}.

We highlight that, importantly, the proposed pretraining is very well-aligned with the downstream task, i.e. supervised class-aware object detection, and it allows the pretraining of \textit{both} the backbone and the detection head simultaneously. This is unlike other detector pretraining methods~\cite{dai2021up,bar2022detreg,joindet} that require freezing the backbone to avoid performance degradation.

\section{Experimental Setting} 
\label{sec:experimental_setting}

We apply \ours to two DETR-based architectures (Deformable DETR~\cite{def_detr} and ViDT+ \cite{vidt}) and two R-CNN architectures (Mask R-CNN~\cite{he2017mask}, Cascade Mask R-CNN~\cite{cai2018cascade}), focusing on the former, as DETR's end-to-end single-stage architecture performs better and is better suited for representation learning.
Following~\cite{bar2022detreg,wang2023cut}, Def. DETR and Cascade Mask R-CNN detectors use ResNet-50~\cite{he2016deep} backbones initialized with SwAV~\cite{swav} and DINO~\cite{dino} respectively. ViDT+ uses a Swin-T~\cite{liu2021swin} backbone initialized with MoBY~\cite{xie2021self}, unless stated otherwise. In all cases, the backbones were trained in a fully unsupervised manner on ImageNet.
To compare with prior work on object detection pretraining, we follow~\cite{bar2022detreg} for Def. DETR and ViDT+,~\cite{wang2023cut} for Cascade Mask R-CNN and~\cite{li2023aligndet} for Mask R-CNN in terms of datasets, hyperparameters, and experiments. 
For unsupervised representation learning, in the absence of a predefined protocol, we use ViDT+ and experiment with the most well-established datasets in object detection. 
Specific hyperparameters and information on the datasets are provided in~\cref{sec:hyperparams} and~\cref{sec:datasets} respectively.
Unless stated otherwise, for methods other than \ours we report results from the respective papers, except where ViDT+ is used.

\section{Experiments}
\label{sec:experimental_results}

We highlight two main results, namely state-of-the-art results for detection pretraining and competitive results for self-supervised representation learning for detection, including pretraining on scene-centric data such as COCO and OpenImages \textbf{from scratch}. We complement these results with a comprehensive set of ablation studies presented in~\cref{sec:ablations}.

\subsection{Object detection pretraining}
\label{sec:od_pretrain}

We evaluate~\ours following the standard protocol for object detection pretraining, as defined by~\cite{bar2022detreg} for DETR-based architectures,~\cite{wang2023cut} for Cascade Mask R-CNN and~\cite{li2023aligndet} for Mask R-CNN, which include experiments in the full-data, semi-supervised and few-shot settings.

\begin{table}[t]
    \caption{\textbf{Object detection results on COCO.} Methods are pretrained on ImageNet, finetuned on MS COCO {\tt train2017} and evaluated on {\tt val2017}. 1: Backbone initialized with MoBY and pretrained with \ours (pretrained detection head was discarded). 
  }\label{tab:main}
  \begin{center}
  \scriptsize{
  \begin{tabular}{ c c c c c }
    \toprule
     \multirow{2}{*}{Detector} & Backbone & Detector & Frozen & \multirow{2}{*}{AP}  \\
     &  Pretraining & Pretraining & Backbone & \\
     \midrule
     \arrayrulecolor{lightgray}
      \multirow{3}{*}{\begin{tabular}{c} Cascade \\ \linebreak Mask R-CNN~\cite{cai2018cascade} \end{tabular}} & \multirow{3}{*}{DINO} & - & \xmark & 44.4 \\
       &  & CutLER~\cite{wang2023cut} & \xmark & 44.7 \\
       &  & \textbf{\ours} & \xmark & \textbf{45.0} \\
       \midrule
     \multirow{7}{*}{Def. DETR~\cite{def_detr}} & \multirow{7}{*}{SwAV} & - & - & 45.2 \\
     &  & UP-DETR~\cite{dai2021up} & \cmark & 44.7 \\
      &  & DETReg~\cite{bar2022detreg} & \cmark & 45.5 \\
      &  & JoinDet~\cite{joindet} & \cmark & 45.6 \\
      &  & SeqCo-DETR~\cite{jin2023seqco} & \cmark & 45.8 \\
      &  & Siamese DETR~\cite{huang2023siamese} & \cmark & 46.3 \\
      &  & \textbf{\ours} & \xmark & \textbf{46.7} \\
     \midrule
     \multirow{5}{*}{ViDT+~\cite{vidt}} & MoBY & - & - & 48.3 \\
     & \textbf{\ours}$^{1}$ & - & - & \textbf{48.8} \\
     & MoBY & DETReg & \cmark & 49.1 \\
     & MoBY & DETReg & \xmark & 47.8 \\
     & MoBY & \textbf{\ours} & \xmark & \textbf{49.6} \\
     \arrayrulecolor{black}
    \bottomrule
  \end{tabular}
  }
  \end{center}
    \vspace{-0.1cm}
\end{table}

\begin{table}[t]
\parbox[b]{.48\linewidth}{
\begin{center}
    \caption{\textbf{Object detection results on PASCAL VOC}. Methods are pretrained on ImageNet, finetuned on PASCAL VOC {\tt trainval07+2012} and evaluated on {\tt test07}.
  }
  \label{tab:pascal}
  \scriptsize{
  \begin{tabular}{ c c c c }
    \toprule
    Method & $AP$ & $AP_{50}$ & $AP_{75}$ \\
    \midrule
    SwAV & 61.0 & 83.0 & 68.1 \\
    DETReg & 63.5 & 83.3 & 70.3 \\
    JoinDet & 63.7 & 83.8 & 70.7 \\
    SeqCo-DETR & 64.1 & 83.3 & 70.3 \\
    \textbf{\ours} & \textbf{64.8} & \textbf{84.6} & \textbf{72.7} \\
    \bottomrule
  \end{tabular}
  }
  \end{center}
  }
\hfill
\parbox[b]{.48\linewidth}{
  \begin{center}
    \caption{\textbf{Object detection results with COCO pretraining}. Mask R-CNN~\cite{he2017mask} is pretrained and finetuned on MS COCO {\tt train2017} and evaluated on {\tt val2017}. The ResNet50 backbone is initialized with SwAV~\cite{swav}.
  }
  \label{tab:aligndet}
  \scriptsize{
  \begin{tabular}{ c c c c }
    \toprule
    Detector & \multirow{2}{*}{$AP$} & \multirow{2}{*}{$AP_{50}$} & \multirow{2}{*}{$AP_{75}$} \\
    Pretraining &  &  &  \\
    \midrule
    - & 41.6 & 62.2 & 45.8 \\
    AlignDet~\cite{li2023aligndet} & 42.3 & 62.5 & 46.7 \\
    \textbf{\ours} & \textbf{43.2} & \textbf{64.2} & \textbf{47.4} \\
    \bottomrule
  \end{tabular}
  }
  \end{center}
  }
  \vspace{-0.1cm}
\end{table}

\setlength{\tabcolsep}{10pt}

\begin{table}[t]
    \caption{\textbf{Semi-supervised results against detector pretraining methods.} 
    Def. DETR is pretrained on MS COCO {\tt train2017}, finetuned on k\% labeled samples, and evaluated on {\tt val2017}.
    }
  \label{tab:semi_sup}
  \begin{center}
  \scriptsize{
  \begin{tabular}{ c c c c c }
    \toprule
    \multirow{2}{*}{Method} & \multicolumn{4}{c}{AP} \\
    \cline{2-5}
    & 1\% & 2\% & 5\% & 10\% \\
    \midrule
    SwAV & 11.79$\pm$0.3 & 16.02$\pm$0.4 & 22.81$\pm$0.3 & 27.79$\pm$0.2 \\
    DETReg & 14.58$\pm$0.3 & 18.69$\pm$0.2 & 24.80$\pm$0.2 & 29.12$\pm$0.2 \\
    JoinDet & 15.89$\pm$0.2 & - & - & 30.87$\pm$0.1 \\
    \textbf{\ours} & \textbf{18.19}$\pm$\textbf{0.1} & \textbf{21.80}$\pm$\textbf{0.2} & \textbf{26.90}$\pm$\textbf{0.2} & \textbf{30.97}$\pm$\textbf{0.2} \\
    \bottomrule
  \end{tabular}
  }
  \end{center}
  \vspace{-0.1cm}
\end{table}

\setlength{\tabcolsep}{5pt}

\begin{table}[t]
    \caption{\textbf{Semi-supervised results against unsupervised localization methods}. FreeSOLO uses SOLOv2~\cite{wang2020solov2} and is pretrained on MS COCO {\tt train2017+unlabeled2017}. CutLER and \ours use Cascade Mask R-CNN and are pretrained on ImageNet. All methods are finetuned on MS COCO {\tt train2017} and evaluated on {\tt val2017}.}
  \label{tab:semi_sup_rcnn}
  \begin{center}
  \scriptsize{
  \begin{tabular}{ c c c c c c }
    \toprule
    \multirow{2}{*}{Method} & \multicolumn{5}{c}{AP (Box / Mask)} \\
    \cline{2-6}
     & 1\% & 2\% & 5\% & 10\% & 100\% \\
    \midrule
    FreeSOLO & - / - & - / - & - / 22.0 & - / 25.6 & - / - \\
    CutLER & 16.8 / 14.6 & 21.6 / 18.9 & 27.8 / 24.3 & 32.2 / 28.1 & 44.7 / 38.5 \\
    \textbf{\ours} & \textbf{20.8} / \textbf{17.5} & \textbf{25.2} / \textbf{21.2} & \textbf{30.0} / \textbf{25.5} & \textbf{33.8} / \textbf{29.0} & \textbf{45.0} / \textbf{38.8} \\
    \bottomrule
  \end{tabular}
  }
  \end{center}
  \vspace{-0.1cm}
\end{table}

\setlength{\tabcolsep}{8pt}

\begin{table}[t]
    \caption{\textbf{Few-shot results}. Def. DETR is pretrained on ImageNet and finetuned on MS COCO {\tt train2014} with $k\in \{10, 30\}$ instances per class. Results reported on the novel classes of {\tt val2014}. DETReg results reproduced in our codebase using the official checkpoint.}
  \label{tab:few_shot}
  \begin{center}
  \scriptsize{
  \begin{tabular}{ c c c c c c c}
    \toprule
    \multirow{2}{*}{Method} & Base Class & \multicolumn{2}{c}{Novel Class AP} & \multicolumn{2}{c}{Novel Class AP$_{75}$} \\
    \cline{3-6}
     & Finetuning & 10 & 30 & 10 & 30 \\
     \midrule
     DETReg & \multirow{2}{*}{\xmark} & 5.6 & 10.3 & 6.0 & 10.9 \\
     \textbf{\ours} &  & \textbf{10.3} & \textbf{14.5} & \textbf{10.9} & \textbf{15.1} \\
     \arrayrulecolor{lightgray}
      \midrule
     DETReg & \multirow{2}{*}{\cmark} & 9.9 & 15.3 & 10.9 & 16.4 \\
     \textbf{\ours} &  & \textbf{12.4} & \textbf{18.9} & \textbf{13.1} & \textbf{20.4} \\
     \arrayrulecolor{black}
    \bottomrule
  \end{tabular}
  }
  \end{center}
  \vspace{-0.1cm}
\end{table}

\setlength{\tabcolsep}{10pt}

\begin{table}[t]
    \caption{\textbf{Object-centric vs Scene-centric pretraining.} \ours is pretrained different datasets, finetuned on MS COCO {\tt train2017} and evaluated on {\tt val2017}.
  }
  \label{tab:open_images}
  \begin{center}
  \scriptsize{
  \begin{tabular}{ c c c c c c}
    \toprule
     Detector Pretraining & AP & AP$_{50}$ & AP$_{75}$ & AR$^{100}$ \\
      \midrule
      - & 48.3 & 66.9 & 52.4 & - \\
      \arrayrulecolor{lightgray}
      COCO & 49.1 & 67.8 & 53.1 & 25.1 \\
      ImageNet &  \textbf{49.6} & \textbf{68.2} & 53.8 & 27.1 \\
      Open Images & 49.4 & 67.9 & \textbf{53.9} & 25.5 \\
      \arrayrulecolor{black}
    \bottomrule
  \end{tabular}
  }
  \end{center}
  \vspace{-0.1cm}
\end{table}

\begin{table}[t]
\setlength{\tabcolsep}{2pt}
\parbox[b]{.63\linewidth}{
\begin{center}
    \caption{\textbf{Pretraining from scratch}.
  \ours is pretrained \textit{without backbone initialization}, finetuned on MS COCO {\tt train2017} and evaluated on {\tt val2017}. For comparison, ViDT+ is finetuned with a MoBY backbone without pretraining.}
  \label{tab:from_scratch}
  \scriptsize{
  \begin{tabular}{ c c c c c c }
    \toprule
     Backbone & Detector & \multirow{2}{*}{Detector} & Pretraining & \multirow{2}{*}{AP} \\
     Pretraining & Pretraining &  & Dataset &  \\
     \midrule
     MoBY & - & \multirow{3}{*}{FCOS*~\cite{tian2019fcos}} & ImageNet & 47.6 \\
     DetCon ~\cite{detcon} & - &  & ImageNet & 48.4 \\
     Odin ~\cite{odin} & - &  & ImageNet & 48.5 \\
     \arrayrulecolor{lightgray}
      \midrule
     - & - & \multirow{5}{*}{ViDT+} & ImageNet & 38.5 \\
      MoBY & - &  & ImageNet & 48.3 \\
     - & \textbf{\ours} &  & COCO & 48.3 \\
     - & \textbf{\ours} &  & Open Images & 48.8 \\
     - & \textbf{\ours} &  & ImageNet & \textbf{49.2} \\
     \arrayrulecolor{black}
    \bottomrule
  \end{tabular}
  }
  \end{center}
    }
\hfill
\setlength{\tabcolsep}{10pt}
\parbox[b]{.33\linewidth}{
  \begin{center}
    \caption{\textbf{Linear probing}. We pretrain \ours with ViDT+ on MS COCO {\tt train2017}, and apply the backbone to linear classification on ImageNet. Results for other methods are taken from~\cite{van2021revisiting}.
  }
  \label{tab:linear_eval}
  \scriptsize{
  \begin{tabular}{ c c }
    \toprule
     Backbone Pretraining & Acc \\
     \midrule
     DenseCL~\cite{densecl} & 49.9 \\
     VirTex~\cite{desai2021virtex} & 53.8 \\
     MoCo~\cite{moco} & 49.8 \\
     Van Gansbeke et al.~\cite{van2021revisiting} & 56.1 \\
     \textbf{\ours} & \textbf{56.4} \\
    \bottomrule
  \end{tabular}
  }
  \end{center}
  }
  \vspace{-0.1cm}
\end{table}

\textbf{Full data setting:} We provide a comprehensive set of comparisons with detector pretraining methods in~\cref{tab:main}, where we pretrain 3 detector architectures on ImageNet, finetune on COCO {\tt train2017} and evaluate on {\tt val2017}. We also report results for ImageNet pretraining and PASCAL VOC finetuning with Def. DETR in~\cref{tab:pascal}. Finally, following the experimental regime proposed by~\cite{li2023aligndet}, we pretrain and evaluate a Mask R-CNN detector on MS-COCO, see~\cref{tab:aligndet}. As~\cref{tab:main,tab:pascal,tab:aligndet} show, our method significantly outperforms competing detector pretraining methods across datasets and with all 4 detector architectures.
Interestingly, all prior work on DETR pretraining requires freezing the backbone. We quantitatively assess the impact of this requirement by making the DETReg backbone trainable, and observe steep performance degradation. Contrary to all these works, \ours supports a trainable backbone due to its better alignment of the pretraining and downstream tasks.

\textbf{Semi-supervised setting:} We present results in~\cref{tab:semi_sup} for Def. DETR, pretrained on COCO {\tt train2017} and fine-tuned on k\% labeled samples, following~\cite{bar2022detreg}. In~\cref{tab:semi_sup_rcnn} we compare with works focusing on unsupervised localization following~\cite{wang2023cut}, where we pretrained a Cascade Mask R-CNN on ImageNet and fine-tuned on COCO {\tt train2017} with k\% samples, including instance segmentation results. 
In both cases, \ours outperforms previous works by large margins, particularly in the more challenging settings with fewer labeled samples. Notably, despite our pretraining being focused on detection, our method outperforms FreeSOLO and CutLER in segmentation performance as well, which highlights its effectiveness.

\textbf{Few-shot setting:} We follow the protocol defined in~\cite{bar2022detreg}, namely we pretrain Def. DETR on ImageNet and report results for two settings on COCO {\tt train2014}: a) we directly finetune on COCO with $k\in \{10, 30\}$ instances from all classes, b) before the k-shot finetuning stage, we first finetune with annotations from the 60 base classes. Results are reported in~\cref{tab:few_shot} on the novel classes of {\tt val2014}, and demonstrate that \ours outperforms DETReg by significant margins. Furthermore, \ours's performance without base class finetuning is very close to its performance with it. These results support that a) our method drastically reduces detector architectures' dependency on annotated data, and b) \ours's learned representations are already class-aware, and the pseudo-labels produced by our method are good enough that \ours can align with COCO's classes with minimal (10-shot) supervision. We conduct a more in-depth analysis of the few-shot setting outcomes and the convergence properties of \ours in~\cref{sec:fsanalysis}.

\subsection{Self-supervised representation learning on scene-centric images} 
\label{sec:ssl_experiments}

In this section, we examine \ours's performance on scene-centric data, and its ability to learn self-supervised representations (i.e., train a backbone). We begin by validating that \ours, when trained on scene-centric data (e.g. COCO), can perform competitively compared to ImageNet pretraining. Then, we use \ours directly for self-supervised representation learning on scene-centric data (i.e., training from scratch on COCO/Open Images), showing promising results. Finally, we show that pretraining on COCO leads to representations that transfer to ImageNet under the linear-probe setting.

\textbf{Object vs Scene-centric pretraining:} In~\cref{tab:open_images}, we present results for \ours when the detector is pretrained on object-centric and scene-centric data of varying quantity. Specifically, keeping the initialization as described in~\cref{sec:experimental_setting}, we further pretrain ViDT+ on MS COCO {\tt train2017} and Open Images.
We additionally report class-unaware object localization performance in terms of Average Recall (AR).
In all cases, our method improves over the baseline, including when we pretrain and finetune on the same dataset (COCO).
We observe that ImageNet performs best, followed by Open Images and MS COCO, though we note that the margins between them are not very large. 
Combined, these findings show that: a) \ours is highly sample efficient, achieving competitive performance even when pretraining with much more limited data (COCO), b) \ours is flexible, being able to handle both object-centric and scene-centric data, and c) the properties of the pretraining dataset are impactful, with a larger object-centric dataset (i.e. ImageNet) leading to better performance relative to scene-centric (Open Images) and smaller (COCO) datasets.
\cref{tab:open_images} provides further insight as to why ImageNet pretraining performs best. As seen by contrasting AR scores, ImageNet's detector localizes more objects correctly. This indicates that the proposals generated for ImageNet are relatively better, which likely leads to better supervision, especially for self-training.
Overall, these results indicate that \ours does not require carefully curated object-centric data although both the size of the dataset and the level of curation have an impact on its performance.

\textbf{Self-supervised representation learning from scratch:} Experiments conducted in previous sections initialize the backbone with weights obtained by self-supervised training on ImageNet. In this section, we evaluate the representation learning capacity of \ours by pretraining a ViDT+ detector from an \textit{untrained} backbone (from scratch) to examine whether independent backbone pretraining is indeed necessary.
We pretrain on object-centric (ImageNet) and scene-centric (COCO \& Open Images) data and present results in~\cref{tab:from_scratch}. For completeness, we include results for other methods that focus on self-supervised backbone-only pretraining, noting that they use a different detector architecture during finetuning.
Results again show that \ours performs best with a well-curated, object-centric pretraining dataset, but is competitive even when trained on complex, scene-centric images. \ours performs on par with backbone-only ImageNet pretraining (MoBY) when pretrained on COCO, and outperforms it when pretrained on Open Images. This outcome supports our thesis that unsupervised pretraining directly on scene-centric data with an object detection task is feasible and effective.

We further evaluate the quality of the COCO-pretrained backbone by performing a linear probe experiment on ImageNet. \cref{tab:linear_eval} shows \ours's performance as well as that of prior work. We note that prior work use a ResNet50 encoder, and thus a direct comparison is hard. It is however clear that our method is competitive, despite being pretrained for object detection, highlighting the natural fit of \ours for general-purpose representation learning from scene-centric images.

\section{Discussion}\label{sec:discussion}

\textbf{Performance:} \ours consistently outperforms previous works in unsupervised object detector pretraining by significant margins. Notably, in the main benchmark of this task (\cref{tab:main}), relative to the baselines, \ours increases the impact of pretraining over the previous state of the art by 100\% for Cascade Mask R-CNN (+0.3 vs +0.6), 36\% for Def. DETR (+1.1 vs +1.5) and 62\% for VIDT+ (+0.8 vs +1.3). Furthermore, we expand on the typical benchmarks of detector pretraining and evaluate \ours when pretraining entirely from scratch (\cref{tab:from_scratch}), showing competitive results. Finally, we examine the impact of the pretraining dataset in terms of its size and level of curation (\cref{tab:open_images}).

\textbf{Complexity:} A major strength of \ours is its simplicity and efficiency. Unlike previous works that draw from contrastive learning, \ours follows the typical pipeline of object detection training without complex auxiliary objectives. Furthermore, previous works~\cite{bar2022detreg,joindet,jin2023seqco,huang2023siamese} use student-teacher architectures during training, which require multiple forward passes per sample and lead to complex pipelines. \ours, on the other hand, requires only one forward pass, making it much more efficient in terms of training speed and memory requirements.

\textbf{Unsupervised pretraining for object detectors:} As mentioned in the introduction, previous works on unsupervised object detector pretraining largely focused on techniques developed for backbone-only self-supervised representation learning, and on adapting them to the detector pretraining task. \ours represents a deviation from this approach, and achieves state-of-the-art results with a simple detection framework, relying on techniques developed for object detection. While \ours is not orthogonal to previous works and could incorporate auxiliary objectives (e.g. based on contrastive learning), we hope our work will motivate research for the development of novel methods in this area that draw from object detection literature, in addition to self-supervised representation learning for backbone architectures.

\textbf{Limitations:} As stated previously, our work is, to the best of our knowledge, the first to consider unsupervised object detection as a pretext task for from-scratch pretraining of detector architectures, as an alternative to the two-stage scheme of backbone-only self-supervised pretraining followed by detector pretraining. Although \ours achieves impressive results in this task, it still requires a pretrained self-supervised model for region proposal and pseudo-label extraction.

\section{Conclusion}

We have proposed \ours, a novel method for self-supervised end-to-end object detector pretraining. Compared to prior work, our method aligns pretraining and downstream tasks through the careful construction of object proposals and pseudo-labels and the use of self-training. We extensively evaluate \ours in several object detector pretraining benchmarks and demonstrate that it consistently outperforms previous methods across settings and detector architectures. However, unlike prior work, we show that \ours is also capable of effectively pretraining the backbone. This brings our method in line with the wider literature on self-supervised representation learning for detection. We again show competitive performance in this area and explore novel settings, specifically pretraining with scene-centric datasets and even pretraining from scratch. Overall, we believe our framework not only outperforms existing detector pretraining methods but also represents a promising step toward self-supervised, fully end-to-end object detection pretraining on uncurated images.

\bibliographystyle{unsrtnat}
\bibliography{main}

\appendix

\newpage

\section{Training Hyperparameters} \label{sec:hyperparams}

We provide here detailed hyperparameters for all training settings included in the main paper. Note however that, for pre-training and fine-tuning, we follow the recipes of the original works. We therefore \textbf{perform no hyperparameter tuning}.
This highlights a positive feature of \ours: transferability. As \ours's training objective is straightforward (pseudo-)class aware detection, which is already what detectors are typically designed for, it can be adapted to any detector architecture with minimal effort.
The specific hyperparameters for each architecture are presented below:\\

\noindent \textbf{Def. DETR:} We follow~\cite{bar2022detreg} and pretrain for 5 epochs per stage on ImageNet with a batch size of 192 and a fixed learning rate of 0.0002. For finetuning, we train on COCO for 50 epochs and PASCAL VOC for 100 epochs, with a batch size of 32. The learning rate is set to 0.0002, and is decreased by a factor of 10 at epoch 40 and 100 for COCO and PASCAL VOC respectively.

\noindent \textbf{ViDT+:} We use the training hyperparameters proposed in~\cite{vidt}. Specifically, unless stated otherwise, ViDT+ is pretrained for 10 epochs per stage on ImageNet and Open Images, and for 50 epochs per stage on COCO, with batch size 128.
In all cases, the learning rate is set to 0.0001 and follows a cosine decay schedule.

\noindent \textbf{Cascade Mask R-CNN:} We use the pretraining and fine-tuning hyperparameters proposed in~\cite{wang2023cut}. Specifically, unless stated otherwise, Cascade Mask R-CNN is pretrained for 160,000 steps per stage on ImageNet with batch size 16. The learning rate is set to 0.01 and decreased by a factor of 10 at after 80,000 training steps.

\noindent \textbf{Mask R-CNN:} We use the pretraining and fine-tuning hyperparameters proposed in~\cite{wang2023cut}. 
This experiment is designed to enable a direct comparison with AlignDet~\cite{li2023aligndet} \textit{under the AlignDet setting}. We note however that it involves limited pretraining (12 epochs) on a small dataset (MS COCO) with a shallow detector architecture (Mask R-CNN), each of these being a disadvantageous regime for \ours, which was designed for pretraining on large datasets with the goal of learning strong representations through object detection.
Accordingly, we modify our approach \textbf{only for this setting} by
a) freezing the backbone (due to the combination of shallow detector, low data regime, and short schedule), and b) avoiding self-training (to restrict training to 12 epochs to match AlignDet).
Even with this setting, which plays against the strengths of \ours, we outperform AlignDet by a significant margin, demonstrating the effectiveness of our proposed aligned pretext task.

Unless stated otherwise, we pretrain with 2048 pseudo-classes (i.e. we set the number of clusters for the global clustering step to 2048), and apply one round of self-training, following our findings in~\cref{tab:abl_stages}.

For experiments other than the full setting (i.e. semi-supervised, few-shot, and self-supervised representation learning), we apply the following changes to the training parameters described above:

\noindent \textbf{Semi-supervised:} For Def. DETR we follow DETReg~\cite{bar2022detreg} and finetune on COCO for 2,000 epochs for 1\% of samples annotated, 1,000 epochs for 2\% of samples, 500 epochs for 5\% of samples, and 400 epochs for 10\% of samples. The learning rate is kept fixed at 0.0002. Results in Table 3 are measured over 5 runs, with different, randomly sampled annotated samples. For Cascade Mask R-CNN, we closely follow the training setting and evaluation protocol used in~\cite{wang2023cut}.

\noindent \textbf{Few-shot:} We finetune on COCO's base classes using the splits proposed in~\cite{wang2020frustratingly}. For the standard few-shot setting we a) finetune on the base classes following the COCO finetuning settings outlined above, and b) finetune on the 10- and 30-shot sets for 30 and 50 epochs respectively, with a fixed learning rate of 0.0002 and 0.00004. For the extreme setting, we directly finetune on the 10- and 30-shot sets for 400 epochs with a learning rate of 0.0002 which is decreased by a factor of 10 after 320 epochs. Results in Table 4 correspond to the best validation score of each run during training, averaged over 5 runs, with k-shot samples corresponding to seeds 1-5 of~\cite{wang2020frustratingly}. When finetuning on the k-shot instances, the backbone is kept frozen in both settings.

\noindent \textbf{Self-supervised representation learning on scene-centric data:} For these experiments, where the entire architecture is initialized from scratch (backbone \& detector), we train for 1,000 epochs on COCO, 100 epochs on ImageNet, and 70 epochs on Open Images. This allows for a fair comparison, with approximately the same number of training steps across datasets.

\section{Datasets} \label{sec:datasets}

We use the training sets of ImageNet~\cite{imagenet2015}, Open Images~\cite{OpenImages2} and MS COCO~\cite{coco2014} for unsupervised pretraining. For supervised finetuning we use the training sets of MS COCO and PASCAL VOC~\cite{everingham2010pascal}. Results are reported for the corresponding validation sets, using Average Precision (AP) and Average Recall (AR).
ImageNet includes 1.2M object-centric images, classified with 1,000 labels and without object-level annotations. Open Images includes 1.7M scene-centric images and a total of 14.6M bounding boxes with 600 object classes. COCO is a scene-centric dataset with 120K training images and 5K validation images containing 80 classes. PASCAL VOC is scene-centric and contains 20K images with object annotations covering 21 classes.

\section{Algorithm}

In this section, we present~\ours as an algorithm.

\begin{algorithm}
\footnotesize{
\caption{\ours Pretraining}
\label{alg:method}
\begin{algorithmic}[1]
\Require $\{X_i\}_{i=1}^I$, Net $g=(g_b,g_h)$, initial params. $\Theta_0$
\State \Comment{Unsup. train set gen. (\cref{sec:initialization})}
\For{$i=1:N$}  
\State ${\mathbf{F}_l} \gets g_b(X_i)$
\State $\mathbb{M}_i \gets \bigcup \text{Cluster}( F_l, K )$ \Comment{$\;K \in \mathcal{K}, l \in \mathcal{L}$}
\State $\mathbb{R}_i \gets $ Connected Components($\mathbb{M}_i$) 
\State $\{b_n^i, f_n^i\}_{N(i)} \gets$ Filter$( \mathbb{R}_i )$ 
\EndFor
\State $\{c_n^i\} \gets $ K-Means$( \{f_n^i\}, K=C)$ \Comment{Pseudo-classes}
\State $\mathcal{T}_0 \gets \left\{ X_i, \{(b_n,c_n)\}_{n=1}^{N(i)} \right\}_{i=1}^I$
\State \Comment{Self-training (\cref{sec:self-training})}
\For{$j$ stages}  
\State $g(-; \Theta_{j+1} ) \gets $ Train $(\mathcal{T}_j, g)$  \Comment{Using eq.~\ref{eq:detr_loss}}
\State $\mathcal{T}_{j+1} \gets $ Filter( $\{g(X_i;\Theta_j)\}_{i=1}^{I}$ )
\EndFor
\end{algorithmic}
}
\end{algorithm}

\section{Convergence \& Alignment Analysis}\label{sec:fsanalysis}

In this section, we discuss the convergence and alignment properties of \ours by analyzing the results of the "extreme" few-shot experiments. As discussed in paper Sec. 5, in this setting we pretrain Def. DETR on ImageNet, and then finetune directly on COCO {\tt train2014}, using $k\in \{10, 30\}$ instances from all classes.

\begin{table}[t]
    \caption{Results of "extreme" few-shot training for 50 epochs and 400 epochs.}
  \label{tab:few_shot_ap}
  \begin{center}
  \begin{footnotesize}
  \begin{tabular}{ c c c c c c c}
    \toprule
    \multirow{2}{*}{Method} & \multirow{2}{*}{Epochs} & \multicolumn{2}{c}{Novel Class AP} & \multicolumn{2}{c}{Novel Class AP$_{75}$} \\
     &  & 10 & 30 & 10 & 30 \\
     \midrule
     DETReg & \multirow{2}{*}{50} & 1.9 & 3.4 & 1.8 & 3.52 \\
     \textbf{\ours} &  & \textbf{8.32} & \textbf{13.9} & \textbf{8.06} & \textbf{14.4} \\
      \midrule
     DETReg & \multirow{2}{*}{400} & 5.6 & 10.3 & 6.0 & 10.9 \\
     \textbf{\ours} &  & \textbf{10.3} & \textbf{14.5} & \textbf{10.9} & \textbf{15.1} \\
    \bottomrule
  \end{tabular}
  \end{footnotesize}
  \end{center}
\end{table}

\begin{figure}[t]
\centering
\hspace{-0.1cm}
\begin{tikzpicture}
\begin{axis}[
    width=0.9\textwidth,
    height=0.35\textwidth,
    xlabel={Epochs},
    ylabel={\textbf{AP}},
    legend style={at={(0.85,0.35)},anchor=north},
    ymin=0,
    xmin=0, xmax=400,
    grid=both,
    scaled y ticks=false,
    yticklabel=\pgfkeys{/pgf/number format/.cd,fixed,precision=1,zerofill}\pgfmathprintnumber{\tick},
    ytick={2.5, 5.0, 7.5, 10.0},
    ]
\pgfplotstableread[col sep=comma]{./figures/AptDet_k10.csv}\datatable
\addplot[
    color=blue,
    line width=1pt,
    ]
    table[y=AptDet, x=Steps] {\datatable};
\addlegendentry{\ours}
\addplot[
    color=red,
    line width=1pt,
    ]
    table[y=DETReg, x=Steps] {\datatable};
\addlegendentry{DETReg}
\end{axis}
\end{tikzpicture}
\caption{AP scores on COCO's {\tt val2014} novel classes during finetuning with k=10 instances per class. Results averaged over 5 runs.}
\label{fig:extreme_k10}
\end{figure}

\begin{figure}[t]
\centering
\hspace{-0.1cm}
\begin{tikzpicture}
\begin{axis}[
    width=0.9\textwidth, 
    height=0.35\textwidth,
    xlabel={Epochs},
    ylabel={\textbf{AP}},
    legend style={at={(0.85,0.35)},anchor=north},
    ymin=0,
    xmin=0, xmax=400,
    grid=both,
    scaled y ticks=false,
    yticklabel=\pgfkeys{/pgf/number format/.cd,fixed,precision=1,zerofill}\pgfmathprintnumber{\tick},
    ytick={2.5, 5.0, 7.5, 10.0, 12.5},
    ]
\pgfplotstableread[col sep=comma]{./figures/AptDet_k30.csv}\datatable
\addplot[
    color=blue,
    line width=1pt,
    ]
    table[y=AptDet, x=Steps] {\datatable};
\addlegendentry{\ours}
\addplot[
    color=red,
    line width=1pt,
    ]
    table[y=DETReg, x=Steps] {\datatable};
\addlegendentry{DETReg}
\end{axis}
\end{tikzpicture}
\caption{AP scores on COCO's {\tt val2014} novel classes during finetuning with k=30 instances per class. Results averaged over 5 runs.}
\label{fig:extreme_k30}
\end{figure}

In~\cref{fig:extreme_k10,fig:extreme_k30} we present the AP scores for \ours and DETReg during training, averaged over 5 runs, and measured over the validation set's novel classes. As was noted in paper Sec. 5, \ours outperforms DETReg by large margins. Notably, however, it is also shown to converge much faster. More specifically, in~\cref{tab:few_shot_ap} we present results for 50 epochs of k-shot finetuning against the performance reached after 400 epochs. In both cases, we average the best validation score across 5 runs. We see that, at 50 epochs, \ours has already reached near-peak performance, while DETReg converges at a much slower rate.

This means \ours effectively alleviates the sample inefficiency and slow convergence of DETR architectures and makes our method particularly useful when annotations and/or computational resources are extremely scarce. These results provide further support for our conclusions in paper Sec. 5, namely that \ours is much better aligned with the downstream task, with learned object representations that are well suited for class-aware object detection, so that minimal training and supervision can lead to strong performance. 

\section{Analysis and ablations}
\label{sec:ablations}

Throughout this section we use ViDT+ and, unless stated otherwise, pretrain on ImageNet for 10 epochs per stage.

\paragraph{Impact of object proposals:} We evaluate our object proposal method in two ways: a) we examine how well it localizes objects by computing the Average Recall (AR) score on COCO {\tt val2017} (see~\cref{tab:rpm}), and b) we investigate its impact on \ours by replacing it with Selective Search and present the outcomes (see~\cref{tab:abl_proposals_tr}).

\begin{table}[t]
    \caption{\textbf{Quality of proposals:} AR results on COCO {\tt val2017}. The first section presents results for the initial extraction of object proposals, while the lower two sections present results for proposals generated by detection/segmentation architectures trained on the initial proposals.}
  \label{tab:rpm}
  \begin{center}
  \begin{footnotesize}
  \begin{tabular}{ c c c c }
    \toprule
     Object proposals & Detection Architecture & AR$^{100}$ \\
     \midrule
     Sel. Search & - & 10.9 \\
      \textbf{\ours}-St. 0 & - & \textbf{13.4} \\
      \midrule
     DETReg & \multirow{3}{*}{ViDT+} & 21.5 \\
     \textbf{\ours}-St. 1 &  & 25.9\\
     \textbf{\ours}-St. 2 &  &\textbf{ 27.1} \\
      \midrule
     CutLER & \multirow{3}{*}{Cascade Mask R-CNN} & \textbf{32.7} \\
     \textbf{\ours}-St. 1 &  & 24.5\\
     \textbf{\ours}-St. 2 &  & 24.6 \\
    \bottomrule
  \end{tabular}
  \end{footnotesize}
  \end{center}
\end{table}

\begin{table}[t]
  \begin{center}
      \caption{\textbf{Impact of initial proposals:} AP results on COCO {\tt val2017}, using different initial object proposal methods.}
  \label{tab:abl_proposals_tr}
  \begin{footnotesize}
  \begin{tabular}{ c c c c c }
    \toprule
     Method & Proposals & AP & AP$_{50}$ & AP$_{75}$ \\
     \midrule
     MoBY & - & 48.3 & 66.9 & 52.4 \\
     \midrule
     \ours-St. 1 & \multirow{2}{*}{Sel. Search} & 48.7 & 67.3 & 52.7 \\
     \ours-St. 2 &  & 48.6 & 67.1 & 52.2 \\
     \midrule
     \ours-St. 1 & \multirow{2}{*}{Our Anns.} & 48.9 & 67.4 & 52.9 \\
     \ours-St. 2 &  & \textbf{49.6} & \textbf{68.2} & \textbf{53.8}  \\
    \bottomrule
  \end{tabular}
  \end{footnotesize}
  \end{center}
\end{table}

\begin{table}[t]
  \begin{center}
      \caption{\textbf{Number of classes}. Pretraining and finetuning on COCO, evaluation in terms of training accuracy, AR of the pretrained detector, and AP of the finetuned model. 1 class implies class-unaware pretraining.}
  \label{tab:abl_clusters}
  \begin{footnotesize}
  \begin{tabular}{ c c c c }
    \toprule
     Classes & ACC & AR & AP \\
     \midrule
     1 & - & \textbf{25.2} & 41.2 \\
    256 & \textbf{80.01} & 23.9 & 43.8 \\
    512 & 75.13 & 24.0 & 43.9 \\
    2048 & 53.75 & 23.9 & \textbf{44.1} \\
    \bottomrule
  \end{tabular}
  \end{footnotesize}
  \end{center}
\end{table}

\cref{tab:rpm} includes results both for our initial proposals (noted as \ours-St. 0), and the proposals generated by pretrained detectors. Results show that our approach is superior to Selective Search and that detector pretraining significantly improves over our initial proposals, supporting our decision to self-train. We observe also that our framework leads to better localization results than DETReg. Most interestingly, we observe that CutLER performs better in terms of localization than~\ours, even though~\ours consistently outperforms CutLER in terms of object detection pretraining. This reinforces our claim in paper Sec. 2, that unsupervised localization methods generate annotations and follow training processes that are not necessarily good for detector pretraining.

In~\cref{tab:abl_proposals_tr} we find that using Selective Search proposals, \ours still outperforms the MoBY baseline, but we observe a performance drop relative to our object proposal method. We attribute this to two reasons: a) our method likely produces more discriminative descriptors $f$ by aggregating representations over a mask of semantically related pixels, rather than over a box, which is the case for Selective Search. This, in turn, leads to better pseudo-labels. b) Our proposals are more robust (see~\cref{tab:rpm}), and therefore provide better supervision. In summary, we conclude that \ours is robust to different object proposal methods, but greatly benefits from an appropriate method choice.

\paragraph{Number of classes:} We ablate the number of pseudo-classes produced by the global clustering of object proposals. For this set of experiments, we pretrain and finetune on COCO {\tt train2017} for 25 epochs each. Note this is a simplified (and cheaper) setting for the purpose of ablating.
We find that, during pretraining, increasing the number of clusters/pseudo-labels leads to decreased training accuracy (ACC) and class-unaware AR (measured on the validation set), which is expected, since increasing the number of classes makes the task harder. However, the AP score after finetuning increases, indicating that the pretrained detector is more powerful. Overall, results indicate that our method is fairly robust to the number of pseudo-labels chosen. We do not increase the number of classes beyond 2048 as training becomes less stable and more computationally expensive.

\paragraph{Self-training stages:} We examine the impact of self-training in \cref{tab:abl_stages}, and find that it produces meaningful gains. We explore additional self-training with ViDT+~\cite{vidt}, but observe no benefits, and therefore limit self-training to one round throughout the paper.

\paragraph{Schedule length:} In~\cref{tab:epochs} we examine the impact of a longer training schedule on our method for both training stages by extending training from 10 to 25 epochs per stage. The results show that a longer training schedule can have some benefits, albeit marginal. Interestingly, \cref{tab:epochs} highlights the importance of self-training, as two training stages totaling a combined 20 epochs (10 per stage) clearly outperform a single training round of 25 epochs.

\paragraph{Time \& VRAM requirements:} We present in~\cref{tab:runtime} runtimes and VRAM usage for \ours and DETReg~
\cite{bar2022detreg}, for pretraining on ImageNet using a VIDT+ detector. We note here that DETReg is among the most efficient methods in the relevant literature~\cite{huang2023siamese}. As seen in~\cref{tab:runtime}, \ours requires slightly more GPU memory than DETReg, which is to be expected given that \ours trains the backbone as well as the detector. However, despite training more parameters and using more proposals \ours has the same training time as DETReg. This demonstrates the increased training efficiency of \ours's class-aware detection framework compared with the contrastive student-teacher pipelines utilized by most previous works on this subject~\cite{li2023aligndet,bar2022detreg,huang2023siamese,joindet}.

Regarding the proposal extraction processes, extracting the initial proposals $\mathcal{T}_0$ is a CPU-intensive process that, in our hardware, requires 24 hours for ImageNet. 
The subsequent proposals $\mathcal{T}_1$ used for self-training are acquired through straightforward inference with the pretrained detector, and require approximately 2 hours to extract.

\begin{table}[t]
  \begin{center}
  \footnotesize{
  \begin{tabular}{ c c c}
    \toprule
      & DETReg & AptDet \\
    \midrule
    Training time (h) & 164 & 165 \\
    Avg. proposals per img. & 28 & 36 / 53 \\
    Memory usage per GPU (GB) & 55.6 & 73.6 \\
    \bottomrule
  \end{tabular}
  }
  \end{center}
    \caption{Runtimes on ImageNet with ViDT+ and 8 V100 GPUs. For \ours, we present the avg. proposals per image for both stages (stage 1/stage 2).} \label{tab:runtime}
\end{table}

\begin{table}[t]
    \caption{\textbf{Self-training rounds.} AP results for ViDT+ pretrained with \ours on ImageNet and finetuned on COCO. Avg. proposals per image are measured during training.}
  \label{tab:abl_stages}
  \begin{center}
  \begin{footnotesize}
  \begin{tabular}{ c c c c c }
    \toprule
     Detector & Stage & AP & AP$_{50}$ & AP$_{75}$ \\
     \midrule
      \multirow{3}{*}{ViDT+~\cite{vidt}} & 1 & 48.9 & 67.4 & 52.9 \\
      & \textbf{2} & \textbf{49.6} & \textbf{68.2} & 53.8 \\
      & 3 & 49.6 & 68.0 & \textbf{53.9} \\
      \midrule
      \multirow{2}{*}{Def. DETR~\cite{def_detr}} & 1 & 46.1 & 64.6 & 50.3 \\
      & \textbf{2} & \textbf{46.7} & \textbf{65.4} & \textbf{50.9} \\
    \bottomrule
  \end{tabular}
  \end{footnotesize}
  \end{center}
\end{table}

\begin{table}[t]
\parbox[b]{.45\linewidth}{
  \begin{center}
      \caption{\textbf{Scheduler length.} AP results for varying training epochs. 10 and 25 epoch Stage 2 models are initialized from 10 and 25 epoch Stage 1 models respectively.}
  \label{tab:epochs}
  \begin{footnotesize}
  \begin{tabular}{ c c c c c }
    \toprule
     Stage & Epochs & AP & AP$_{50}$ & AP$_{75}$ \\
     \midrule
      1 & 10 & 48.9 & 67.4 & 52.9 \\
      1 & 25 & 49.2 & 67.7 & 53.6 \\
      \midrule
      2 & 10 & 49.6 & 68.2 & 53.8 \\
      2 & 25 & 49.7 & 68.1 & 54.2 \\
    \bottomrule
  \end{tabular}
  \end{footnotesize}
  \end{center}
  }
\hfill
\parbox[b]{.45\linewidth}{
  \begin{center}
      \caption{\textbf{Data augmentations.} \ours's performance with/without mosaic transforms, pretrained on ImageNet and evaluated on MS COCO.}
  \label{tab:abl_mosaic}
  \begin{footnotesize}
  \begin{tabular}{ c c c}
    \toprule
     Method & Augmentations & AP \\
     \midrule
      CutLER~\cite{wang2023cut} & Copy-paste~\cite{dwibedi2017cut} & 44.7 \\
      \ours & \xmark & 44.8 \\
      \ours & Mosaic~\cite{bochkovskiy2020yolov4} & \textbf{45.0} \\
    \bottomrule
  \end{tabular}
  \end{footnotesize}
  \end{center}
    }
\end{table}

\paragraph{Impact of augmentations:} Unlike most previous works, \ours does not use contrastive learning. Therefore we do not use the color and cropping augmentations that are standard in contrastive learning. Since our task is better aligned with detection, we use mosaic~\cite{bochkovskiy2020yolov4} transforms, a standard object detection augmentation. We evaluate the impact of this choice in~\cref{tab:abl_mosaic} with a Cascade Mask R-CNN detector. While using augmentations helps, we observe that even \textit{without augmentations}, \ours achieves state-of-the-art performance. These findings provide strong evidence that the core performance differentiator is our framework.

\newpage

\section{Visualization} \label{sec:viz}

  \begin{figure}[!htbp]
   \begin{center}
    \includegraphics[width=1.\linewidth]{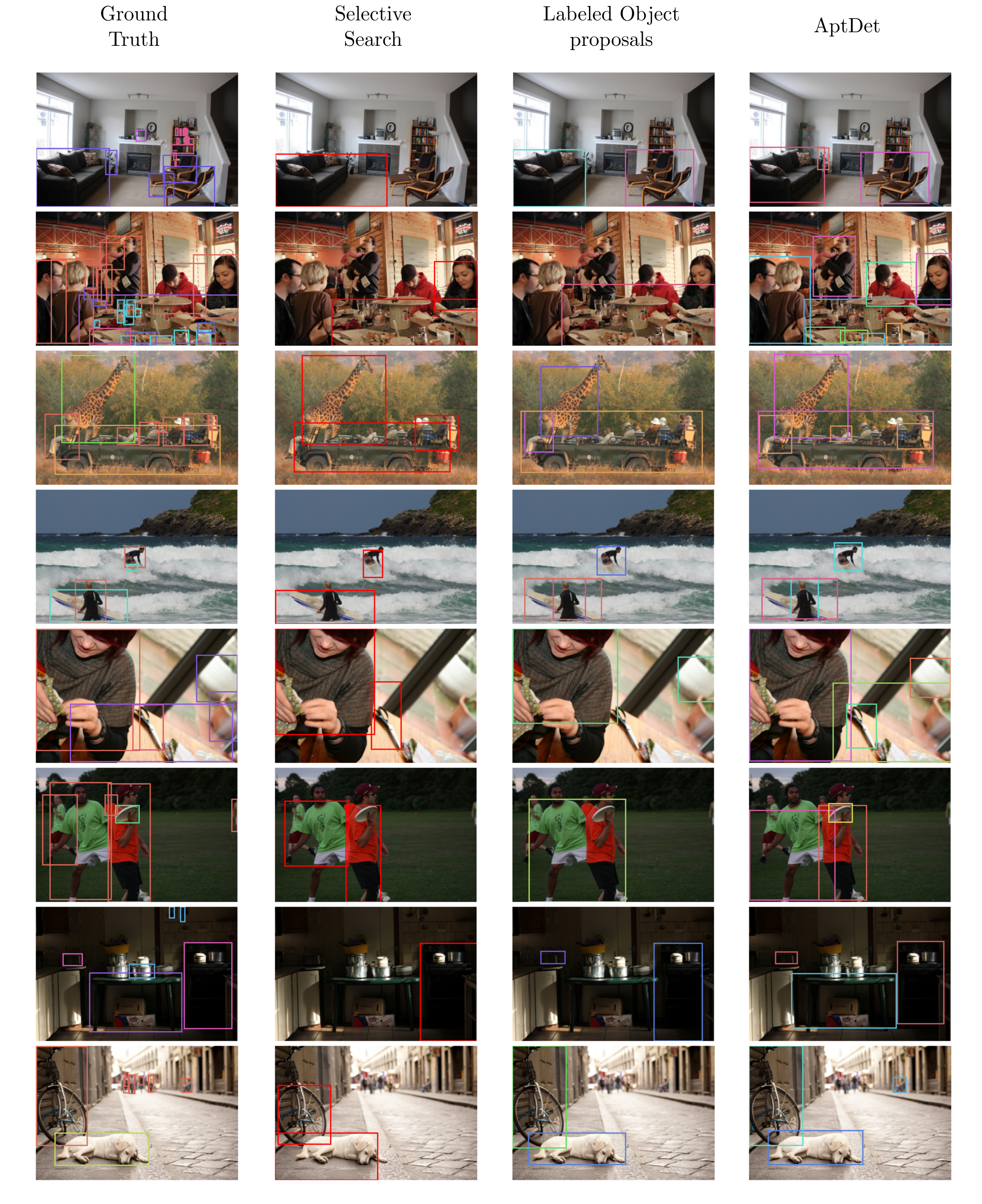}
    \caption{Examples of object proposals extracted from \ours, contrasted with the ground truth, Selective Search and our initial pseudo-labeled object proposals, extracted as described in paper Sec. 3.1. The images belong to COCO {\tt train2017}. To avoid clutter, we only show predicted objects whose bounding boxes have an IOU greater than 0.5 with at least one ground truth object. Best seen in color.}
    \label{fig:viz}
    \end{center}
 \end{figure}

In~\cref{fig:viz} we provide examples visual examples of bounding boxes produced by Selective Search, our pseudo-labeled object proposal method, and \ours, specifically a ViDT+ detector trained for two stages on ImageNet. To avoid clutter, for all three methods we only include objects whose predicted bounding boxes have an IOU higher than 0.5 with an object in the ground truth set.

The images illustrate that self-training significantly improves the object discovery performance of \ours over the original region proposals. Notably, those include much smaller items, and much better performance in cluttered scenes. As stated in the main paper, this contributes to the performance of our framework and specifically the performance gains between stages.

\end{document}